\documentclass[runningheads,a4paper]{llncs}

\usepackage{amssymb}
\usepackage{graphicx}
\usepackage{amsmath}
\usepackage{makecell}
\usepackage{subcaption}
\usepackage{mdframed}
\usepackage[normalem]{ulem}
\usepackage{fontawesome}
\usepackage{todonotes}
\usepackage{booktabs}
\newcommand{\ra}[1]{\renewcommand{\arraystretch}{#1}}
\usepackage{multirow}

\usepackage{algpseudocode}
\usepackage{algorithm}
\usepackage[hidelinks]{hyperref}

\usepackage{url}
\urldef{\mailsa}\path|{2soll, hinz, magg, wermter}@informatik.uni-hamburg.de|
\newcommand{\keywords}[1]{\par\addvspace\baselineskip
\noindent\keywordname\enspace\ignorespaces#1}

\newcommand{\myheader}{Accepted as a conference paper at the \textit{International Conference\\\hskip 3cm on Artificial Neural Networks (ICANN)} 2019}

\usepackage{atbegshi,picture}

\usepackage{datetime}            

\usepackage{lastpage}            

\newcommand{\myleftstd}{1.6in}

\ifthenelse{\isodd{\thepage}}
 {\newcommand{\myleftmargin}{\oddsidemargin+\myleftstd}}
 {\newcommand{\myleftmargin}{\evensidemargin+\myleftstd}}

\AtBeginShipoutNext { 
 \AtBeginShipoutUpperLeft{%
  \put(\dimexpr 0.2\myleftmargin\relax,-1.2cm){\parbox{\textwidth}{\footnotesize\sffamily\noindent{\centering \myheader \hfill}}}%
}}

\title{Evaluating Defensive Distillation For Defending Text Processing Neural Networks Against Adversarial Examples}

\titlerunning{Evaluating Defensive Distillation Against Text Adversarial Examples}

\author{Marcus Soll \and Tobias Hinz \and Sven Magg \and Stefan Wermter}

\authorrunning{M. Soll, T. Hinz, S. Magg, S. Wermter}

\institute{Knowledge Technology, Department of Informatics, Universit\"at Hamburg\\
Vogt-Koelln-Str. 30, 22527 Hamburg, Germany
\url{https://www.inf.uni-hamburg.de/en/inst/ab/wtm}\\
\mailsa}

\toctitle{Evaluating Defensive Distillation For Defending Text Processing Neural Networks Against Adversarial Examples}
\tocauthor{Marcus Soll, Tobias Hinz, Sven Magg, Stefan Wermter}

\begin{document}

\mainmatter  

\maketitle


\begin{abstract}
\emph{
 Adversarial examples are artificially modified input samples which lead to misclassifications, while not being detectable by humans.
 These adversarial examples are a challenge for many tasks such as image and text classification, especially as research shows that many adversarial examples are transferable between different classifiers.
 In this work, we evaluate the performance of a popular defensive strategy for adversarial examples called defensive distillation, which can be successful in hardening neural networks against adversarial examples in the image domain.
 However, instead of applying defensive distillation to networks for image classification, we examine, for the first time, its performance on text classification tasks and also evaluate its effect on the transferability of adversarial text examples.
 Our results indicate that defensive distillation only has a minimal impact on text classifying neural networks and does neither help with increasing their robustness against adversarial examples nor prevent the transferability of adversarial examples between neural networks.
 }
\end{abstract}

\section{Introduction}

 One of the main goals in neural network research is the creation of robust models, especially against noise in the input data. A special form of noise are so-called adversarial examples, first discovered by Szegedy et al. \cite{Szegedy2014}.
 This special type of noise is explicitly crafted to make a classifier misclassify samples without being detectable by humans (up to manipulating the sample so that it is classified to any class the adversary desires).
 This impact is increased for image classification by a property called transferability \cite{Szegedy2014}, which means that adversarial images created for one network have a high chance of being also misclassified by networks with different architectures or training sets.

 Recently, adversarial examples have also been created for deep neural networks used for text classification \cite{Liang2017,Samanta2017,Zhang2019}. Such examples are challenging for a lot of cases, such as automatic indexing and text filtering.
 In automatic indexing, adversarial examples could change the index of a document, e.g. to push an advertising article into a different category, while in text filtering adversarial examples could change the filter outcome, e.g. change a spam e-mail so it is not detected by the spam filter.
 Since deep neural networks now achieve similar or better results compared to traditional methods for text classification (e.g.\ decision trees or support vector machines) \cite{Kim2014,Zhang2015,Zhang2015_2}, adversarial examples can prove problematic for real-world text classification applications.

 Consequently, one of the main research goals with respect to adversarial examples is to develop defense mechanisms that make neural networks less susceptible to these adversarial examples.
 This is especially important for neural networks that are applied in sensitive application areas.
 The goal is, therefore, to harden these neural networks in order to prevent  (adversarial) misclassifications.
 There already exist methods for hardening (deep) neural networks for image classification, such as defensive distillation \cite{Papernot2016_1,Papernot2017}, Batch Adjusted Network Gradients \cite{Rozsa2018}, or detecting adversarial examples like in SafetyNet\cite{Lu2017}.
 However, such work is currently missing for deep neural networks for text classification.

 In this work, we examine whether it is possible to transfer one popular defense mechanism, called defensive distillation \cite{Papernot2016_1,Papernot2017}, for adversarial image examples to the text classification domain in order to increase the robustness of the deep neural network against both adversarial examples and their transferability.
 We use the algorithm of Samanta and Mehta \cite{Samanta2017} for generating adversarial examples in the text domain and evaluate the effectiveness of defensive distillation on two datasets.
 Our experiments show no increased robustness of networks trained with defensive distillation for both, directly generated adversarial examples and adversarial examples generated for a network trained without defensive distillation.
 Additionally, our results show that defensive distillation is also not effective in preventing the transfer of adversarial examples from one network to another.

\section{Related work}
 Szegedy et al. \cite{Szegedy2014} introduced adversarial examples for deep neural networks for image recognition.
 They also added the concept of transferability of adversarial examples between neural networks with different architectures and trained on different datasets.
 The first proposed method for generating adversarial examples of images is the \textit{fast gradient sign method} (FGSM) by Goodfellow et al. \cite{Goodfellow2014}.
 To find these samples, the sign of the cost function's gradient (with respect to the input) is added to the original image.
 Because this basic approach of using the gradient is used by many of the current methods, the gradient has an important role for both generating adversarial examples and hardening networks.

 As a result, many hardening methods perform \textit{gradient masking} \cite{Papernot2016_5}.
 While this can help against white-box attacks (where the process generating the adversarial examples has access to the model), it does not work for methods based on probability \cite{Su2017} or black-box attacks.
 In black box attacks, the generating process has no direct access to the model and needs to generate adversarial examples by other means, e.g., by creating them on a different network and exploiting transferability.
 An overview of the current research on adversarial examples focusing mostly on image classification can be found in a survey by Akhtar and Mian \cite{Akhtar2018}, who summarize different popular generation methods (including real-world examples), the current understanding of adversarial examples, and different methods of increasing robustness of neural networks.

 Since text is discrete rather than continuous (like images), methods for creating adversarial examples for images cannot directly be applied to the text domain \cite{Zhang2019}.
 In addition, small changes can be detected more easily by humans and can easily change the semantics of a sentence \cite{Zhang2019}.
 Liang et al. \cite{Liang2017} were the first to examine adversarial examples for text-processing deep neural networks.
 They focused on so-called \textit{Hot Training Phrases (HTP)} and \textit{Hot Sample Phrases (HSP)}, which are phrases which contribute strongly to the determination of the predicted class.
 Both \textit{HTP} and \textit{HSP} are determined using the cost gradient of the prediction, and both are then used in conjunction for modifying the original sample to generate an adversarial example.
 Through insertion, modification and deletion as well as combining these three strategies, Liang et al. \cite{Liang2017} are able to generate adversarial examples.

 Samanta and Mehta \cite{Samanta2017} extend Liang et al.'s algorithm using the same three basic operations (insertion, modification, deletion). The algorithm works on a current word $w_i$ which changes every round, ordered by the cost gradient (the word with the highest gradient is chosen first):
\begin{enumerate}
 \item If the current word $w_i$ is an adverb, it is deleted (because this operation often does not change the grammar of the sentence).
 \item Else, a word $p_i$ is chosen from a candidate pool $P$ and processed as follows:
 \begin{enumerate}
  \item If the chosen word $p_i$ is an adverb and the current word $w_i$ is an adjective, the chosen word $p_i$ is placed before the current word $w_i$.
  \item Else, the current word $w_i$ is replaced with the chosen word $p_i$.
 \end{enumerate}
\end{enumerate}
 The candidate pool is built from \textit{synonyms}, \textit{typos} and \textit{genre specific keywords} which are words which can only be found in one class.

 This was later followed up by more work as shown in a survey by Zhang et al. \cite{Zhang2019}.
 Besides different ways of generating adversarial examples, they also report two ways of increasing robustness used in the literature: data augmentation (where adversarial examples are included in the training set) and adversarial training (changing the learning function to include adversarial examples as additional regularization).
 However, most current defense mechanisms can be circumvented.
 Data augmentation sometimes helps to prevent adversarial examples generated with the same method as the augmented adversarial examples but fails to harden the network against adversarial examples generated by methods that were not used for the augmented dataset.
  Jia and Liang \cite{Jia2017} present another example where data augmentation does not help against modified generation algorithms.
 As a result, developing effective defense mechanisms against adversarial examples in the text domain remains as a challenge for many real-world applications.

\begin{figure}[tb]
  \centering
  \includegraphics[width=0.75\textwidth]{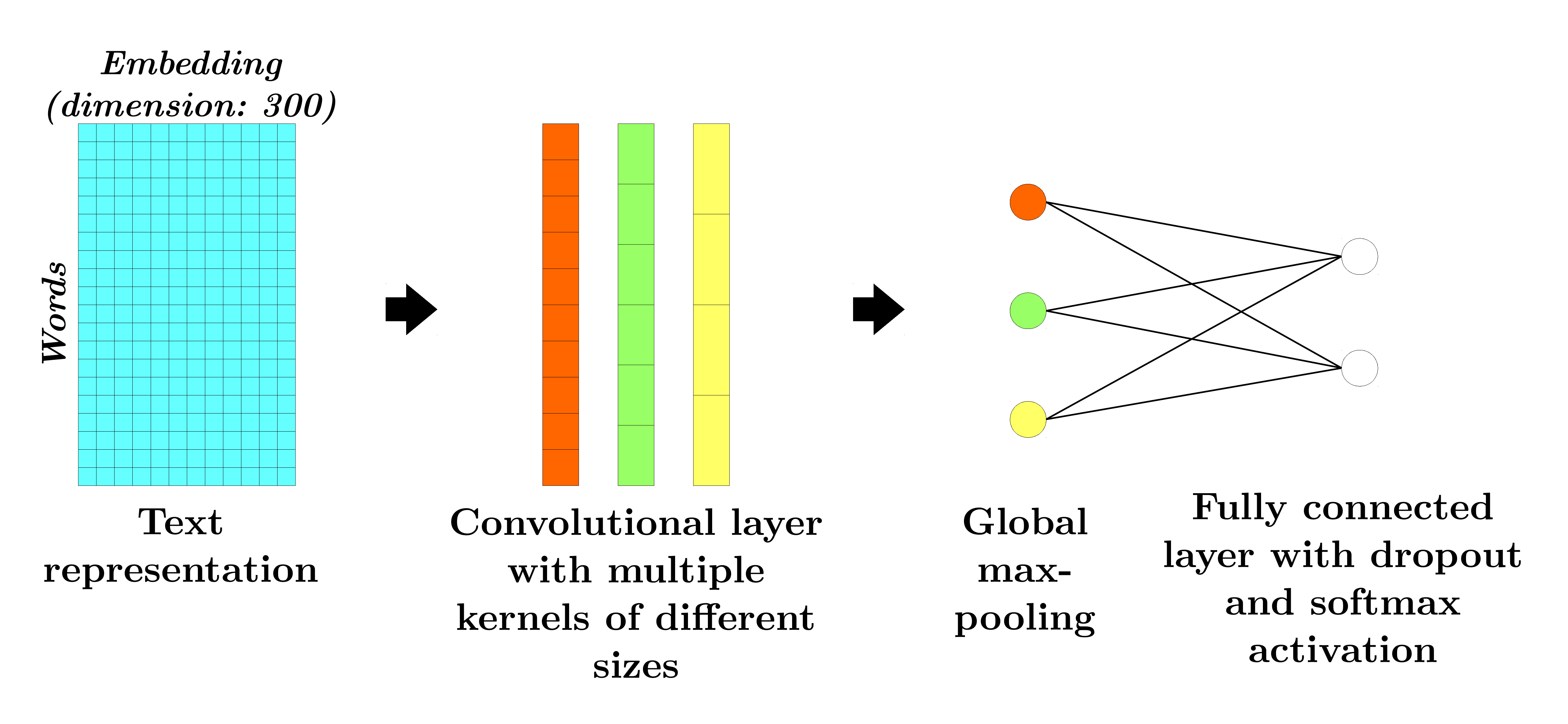}
  \caption{Single-layer convolutional neural network model with different kernel sizes for text classification.}
  \label{fig:nn_architecture}
 \end{figure}
\section{Methodology}

\subsection{Datasets}
 We used the following two datasets as the basis for generating adversarial examples in this paper and to evaluate the effects of defensive distillation:

\vspace{-0.5em}
\paragraph{AG's corpus of news articles:}
 The AG's corpus of news articles is a dataset of news article metadata (including URL, title, category, description, and publication date) \cite{AG}.
 The metadata of almost 500.000 articles can be found in the XML version.
 Given this description, the task is to predict the category of the article.
 Similarly to Zhang et al. \cite{Zhang2015} we only consider the four largest categories: \textit{World}, \textit{Entertainment}, \textit{Sports}, and \textit{Business}.
 Because of the size of the dataset, only the first 4000 articles of each category are used for the training set (total: 16000) and the following 400 (total: 1600) for the test set.

\vspace{-0.5em}
\paragraph{Amazon movie reviews:}
 The Amazon movie reviews dataset \cite{McAuley2013}, taken from the Stanford Network Analysis Project \cite{snapnets}, contains Amazon movie reviews from the years 1997-2002.
 The task is to rate the review as \emph{good} ($\geq 4.0$) or \emph{bad} ($\leq 2.0$) based on the text of the review.
 Because of the size of the dataset, 2000 reviews for each category were taken into consideration (total: 4000), with an additional 200 (total: 400) from each category for the test set.

\subsection{Text encoding}
 In order to process the text with a neural network, we need an appropriate representation of it.
 We chose to encode our words with word2vec \cite{Mikolov2013_2}, which encodes each word into a single vector containing continuous values.
 The resulting vectors are concatenated and used as input for our network.
 In this paper, we used the model trained by Mikolov et al. \cite{Mikolov2013} on the Google news corpus to encode each word into a 300-dimensional vector.

\subsection{Neural Network Model}
 We use a convolutional neural network (CNN) similar to the ones proposed by Kim \cite{Kim2014} and Zhang and Wallace \cite{Zhang2015_2}.
 The architecture (see Fig.~\ref{fig:nn_architecture}) consists of a single convolutional layer with multiple kernels, where the kernels have different window sizes (we use window sizes of $3$, $4$, and $5$ in our network).
 The convolutional layer is followed by a global max-pooling and a fully connected layer with dropout.
 The advantage of this simple architecture is its fast training and high flexibility due to the different kernel sizes.
 All networks are trained using categorical cross-entropy as the loss function, as used by Zhang and Wallace \cite{Zhang2015_2}.

 \begin{figure}[tb]
 \begin{subfigure}[b]{1.0\textwidth}
  \begin{mdframed}[backgroundcolor=white]
  \footnotesize {
   AP - Pau Gasol and Spain won the Olympics' first matchup of NBA stars Sunday, easily beating Yao Ming and China. Gasol, who plays for the Memphis Grizzlies, had 21 points and 10 rebounds despite missing 6 minutes in the first quarter because of a \textcolor{blue}{Yukos} bloody nose, and Spain won 83-58.
   }
  \end{mdframed}
  \vspace{-1em}
  \caption{The network's classification changes from \emph{Sports} (correct) to \emph{World} (incorrect) through the insertion of the word ``\emph{Yukos}'' (highlighted in blue).} 
  \vspace{1em}
 \end{subfigure}
 \begin{subfigure}[b]{1.0\textwidth}
  \begin{mdframed}[backgroundcolor=white]
  \footnotesize {
    Denzel Washington will next direct The Great Debaters, a film based on a true story about an all-black high school debate \textcolor{red}{eventing (team)} in 1935, MTV reports
   }
  \end{mdframed}
  \vspace{-1em}
  \caption{The network's classification changes from \emph{Entertainment} (correct) to \emph{Sports} (incorrect) through the modification of the word ``\emph{team}'' which is replaced by the word ``\emph{eventing}'' (highlighted in red).} 
  \vspace{1em}
 \end{subfigure}
 \vspace{-3em}
 \caption{Generated adversarial examples on the AG dataset.}
 \vspace{-1em}
 \label{fig:generated:examples:AG}
\end{figure}

\subsection{Creating Adversarial Examples}
 Formally, the problem of finding an adversarial example can be defined as following: given a model $f$ (e.g.\ a neural network) and an input $x$ with the label $y$, find some noise $\epsilon$ so that $f(x + \epsilon) = y'$ with $y' \neq y$.
 To avoid detection by humans, the noise $\epsilon$ should be as small as possible.

 In this work, we use a version of the algorithm by Samanta and Mehta \cite{Samanta2017}, where the candidate pool $P$, from which possible words for insertion and replacement are drawn, was created from the following sources:
\begin{itemize}
 \item \textit{Synonyms} gathered from the WordNet dataset \cite{Fellbaum1998},
 \item \textit{Typos} from a dataset \cite{Mitton} to ensure that the typos inserted are not recognized as artificial since they occur in normal texts written by humans, and
 \item \textit{Keywords} specific for one input class which were found by looking at all training sentences and extracting words only found in one class.
\end{itemize}
 Words from the candidate pool were only considered if the part of speech (e.g. \textit{plural noun}) matches the target word.
 Examples of adversarial samples created in this paper can be seen in Fig.~\ref{fig:generated:examples:AG} and Fig.~\ref{fig:generated:examples:amazon}.

\begin{figure}[bt]
  \begin{subfigure}[b]{1.0\textwidth}
   \begin{mdframed}[backgroundcolor=white]
   \footnotesize {
    Still haven't got it. I ordered this series a month ago, I received in the mail a \sout{completely} \textcolor{red}{trivia (different)} movie addressed to someone in connecticut ( I live in Louisiana) I then payed out of my own pocket to mail it to the address on the slip.  I contacted the seller and they said they would work on getting me my product.  Then yesterday I sent an email to see what was going on and they told me they have confirmation of delivery a month ago.  So now I have to reexplain what happened to what I am assuming is a different employee.
    }
   \end{mdframed}
   \vspace{-1em}
   \caption{The network's classification changes from \emph{bad} (correct) to \emph{good} (incorrect) through the deletion of the word ``\emph{completely}'' (crossed out) and the modification of the word ``\emph{different}'' which is replaced by the word ``\emph{trivia}'' (highlighted in red).} 
   \vspace{1em}
 \end{subfigure}
 \begin{subfigure}[b]{1.0\textwidth}
  \begin{mdframed}[backgroundcolor=white]
  \footnotesize {
    Don't buy The quality is horrible. The screen shakes through the whole movie. The title screen \textcolor{red}{excels (is)} all in spanish even though the movie is in english. You would be better off recording it off the television.
   }
  \end{mdframed}
  \vspace{-1em}
  \caption{The network's classification changes from \emph{bad} (correct) to \emph{good} (incorrect) through the modification of the word ``\emph{is}'' which is replaced by the word ``\emph{excels}'' (highlighted in red).} 
  \vspace{1em}
\end{subfigure}
 \begin{subfigure}[b]{1.0\textwidth}
  \begin{mdframed}[backgroundcolor=white]
  \footnotesize {
   Still enjoy watching. Watching older movies, such as this one, can be quite enjoyable. There is no reliance on special effects, just what is needed to get to points. Otherwise this is a people movie commenting on Welles' vision of what may be. Rod Taylor plays his part well and for once was not the cavalier playboy as in others.  A nice piece was the featurette that included an \textcolor{blue}{ONLY} updated visit of Welles and his friend. Nothing fancy, good watching. }
  \end{mdframed}
  \vspace{-1em}
  \caption{The network's classification changes from \emph{good} (correct) to \emph{bad} (incorrect) through the through the insertion of the word ``\emph{ONLY}'' (highlighted in blue).} 
\end{subfigure}
 \vspace{-2em}
 \caption{Generated adversarial examples on the Amazon movie dataset.}
 \label{fig:generated:examples:amazon}
\end{figure}

\subsection{Defensive Distillation}
 Defensive distillation is a method proposed by Papernot et al. \cite{Papernot2016_1,Papernot2017} and is based on the \textit{distillation} method by Hinton et al. \cite{Hinton2015}.
 Both methods are based on the idea that knowledge from one neural network can be transferred to another neural network by using \textit{soft labels} (the output of a previously trained network which represents the probability of the different classes) for the training instead of \textit{hard labels} (where every data belongs to exactly one class).
 To achieve this effectively, the soft labels have to be calculated according to  the following equation:
\begin{equation*}
 y_i = \frac{e^{l_i/T}}{\sum_i e^{l_i/T}},
\end{equation*}
where $y_i$ is the probability of the $i$-th class, $l_i$ the $i$-th \textit{logit} (the inputs to the final softmax level) and $T$ the \textit{temperature}.
 The temperature is used to control how ``soft'' the resulting labels are.
 A high $T$ ($T\rightarrow\infty$) means that a sample has a uniform probability of belonging to any class and a small $T$ ($T\rightarrow 0^+$) means that the label becomes more similar to a one-hot vector.
 A special case is $T=1$, which equals the normal \textit{softmax}.

 These soft labels can now be used to transfer knowledge from the original network to a distilled one.
 The original network is trained as usual and the soft labels are then calculated for the training set using a high temperature (e.g.\ Papernot et al. \cite{Papernot2016_1} suggest a temperature of $T=20$).
 These soft labels are then used to train the distilled network, which has to use the same temperature in its last layer during training.
 After that, the temperature is set back to $T=1$ and the distilled network can be used normally.

 The difference between distillation and defensive distillation is that Hinton et al. \cite{Hinton2015} use distillation to transfer knowledge from a large neural network to a small one while retaining accuracy, whereas Papernot et al. \cite{Papernot2016_1,Papernot2017} use defensive distillation to transfer knowledge from one network to another one with the same size with the goal of making it harder to find adversarial examples for the distilled network.
 In this work, we use the variant described by Papernot et al. \cite{Papernot2016_1} and the network is trained on both the hard and the soft labels since this can significantly improve the process according to Hinton et al. \cite{Hinton2015}.

\section{Experiment setup}
 To examine whether defensive distillation \cite{Papernot2016_1,Papernot2017} has any effects on adversarial examples for text classification, we trained neural networks with and without defensive distillation.
 After that, adversarial examples were generated for each of the networks\footnote{The software used for the experiments can be found online at \url{https://github.com/Top-Ranger/text_adversarial_attack}}.
 Since the goal is to determine whether defensive distillation actually increases the robustness against adversarial examples, it is not necessary to find the optimal hyperparameters for each dataset.
 Therefore, the temperature $T = 20$ chosen for all experiments is the one used by Papernot et al. \cite{Papernot2016_1} for the MNIST dataset.
 To get a better overview over the influence of the temperature, the temperatures $T = 10$, $T = 30$ and $T = 40$ were also tested.
 For the training, both the soft labels and the hard labels were used, where the loss function consists of $10\%$ of the hard label loss and $90\%$ of the soft label loss.
 All networks were trained for 10 epochs since no notable improvement of accuracy occurred after that.

 The question remains whether transferability is an inherent property of adversarial examples or if transferability can be prevented.
 Some recent research \cite{Tramer2017_2} suggests that it might be possible to prevent transferability, however, at the time of writing more research is needed in this direction.
 To test whether defensive distillation has any effect on transferability, we use adversarial examples created on the network trained without distillation and investigate whether these are also misclassified by the neural network trained with defensive distillation.
 An adversarial example is tested on the distilled network if the distilled network predicts the class of the corresponding unaltered input correctly.
 As a baseline, the same examples are tested on retrained networks without distillation.

 \begin{table}[t]
  \centering
  \ra{1.2}
  \caption{\emph{Accuracies} of defensively distilled networks trained with different Temperatures $T$ against test set of the specified dataset.}
  \label{tab:distillation:precision+loss}

  \setlength{\tabcolsep}{1em}
  \begin{tabular}{@{}lccccc@{}}
  \toprule
   Dataset            & $T = 10$ & $T = 20$ & $T = 30$ & $T = 40$ & \begin{tabular}{@{}c@{}} w/o\\distillation \end{tabular} \\
   \midrule
   AG                 & 0.733 & 0.728 & 0.739 & 0.744 & 0.759 \\
   Amazon movie       & 0.798 & 0.825 & 0.795 & 0.780 & 0.885 \\
   \bottomrule
  \end{tabular}
 \end{table}

\begin{table}[b]
 \centering
  \ra{1.2}
 \caption{\emph{Success rates} of generating adversarial examples for networks trained with defensive distillation with different Temperatures $T$.}
 \label{tab:distillation:results}

 \setlength{\tabcolsep}{1em}
 \begin{tabular}{@{}lccccc@{}}
 \toprule
  Dataset & $T = 10$ & $T = 20$ & $T = 30$ & $T = 40$ & \begin{tabular}{@{}c@{}} w/o\\distillation \end{tabular} \\
  \midrule
  AG & 0.976 & 0.981 & 0.972 & 0.978 & 0.982 \\
  Amazon movie & 0.961 & 0.966 & 0.974 & 0.966 & 0.984 \\
  \bottomrule
 \end{tabular}
\end{table}

\section{Results}
 The accuracies of the networks trained with and without defensive distillation are reported in Tab.~\ref{tab:distillation:precision+loss}.
 The performance of the networks trained with distillation is slightly worse compared to the networks trained without distillation, which is expected according to Papernot et al. \cite{Papernot2016_1}.
 The accuracy achieved in the experiment is comparable to others for the Amazon movie dataset (0.82-0.95 by Zhang et al. \cite{Zhang2015}) and slightly lower for the AG dataset (0.83-0.92 by Zhang et al. \cite{Zhang2015}).

 When we look at the success rate of generating adversarial examples (see Tab.~\ref{tab:distillation:results}), the difference between the distilled and the non-distilled networks is marginal at best.
  Overall, the success rate of generating adversarial examples is high with $96\%$ to $98\%$ percent, with no visible difference between the different temperatures.
  This is surprising, since the experiments of Papernot et al. \cite{Papernot2016_1} showed improved robustness for image processing networks even for low temperatures.

\begin{table}[t]
 \centering
  \ra{1.2}
 \caption{\emph{Number of changes} for generating adversarial examples with defensive distillation and Temperature T. Numbers in brackets are without distillation as comparison.}
 \label{tab:distillation:detail}

 \setlength{\tabcolsep}{0.75em}
 \begin{tabular}{@{}llrrcc@{}}
 \toprule
  T & Dataset      & \makecell{Mean length\\ of sentences in\\ successful runs} & \makecell{Mean\\ number \\of changes} & \makecell{Median\\ number \\of changes} & \makecell{Mode\\ number \\of changes} \\
  \midrule
  \multirow{2}{*}{10} & AG & 32.12 (32.76) & 4.47 (3.52) & 2 (2) & 1 (1) \\
   & Amazon movie & 148.85 (139.80) & 19.29 (14.76) & 4 (3) & 1 (1) \\ \midrule
  \multirow{2}{*}{20} & AG & 32.01 (32.76) & 4.18 (3.52) & 2 (2) & 1 (1) \\
   & Amazon movie & 143.05 (139.80) & 18.39 (14.76) & 4 (3) & 1 (1) \\ \midrule
  \multirow{2}{*}{30} & AG & 32.00 (32.76) & 4.13 (3.52) & 2 (2) & 1 (1) \\
   & Amazon movie & 140.04 (139.80) & 18.47 (14.76) & 4 (3) & 1 (1) \\ \midrule
  \multirow{2}{*}{40} & AG & 31.76 (32.76) & 3.94 (3.52) & 2 (2) & 1 (1) \\
   & Amazon movie & 141.24 (139.80) & 17.31 (14.76) & 4 (3) & 1 (1) \\
  \bottomrule
 \end{tabular}
\end{table}

\begin{figure}
  \centering
  \includegraphics[width=0.9\textwidth]{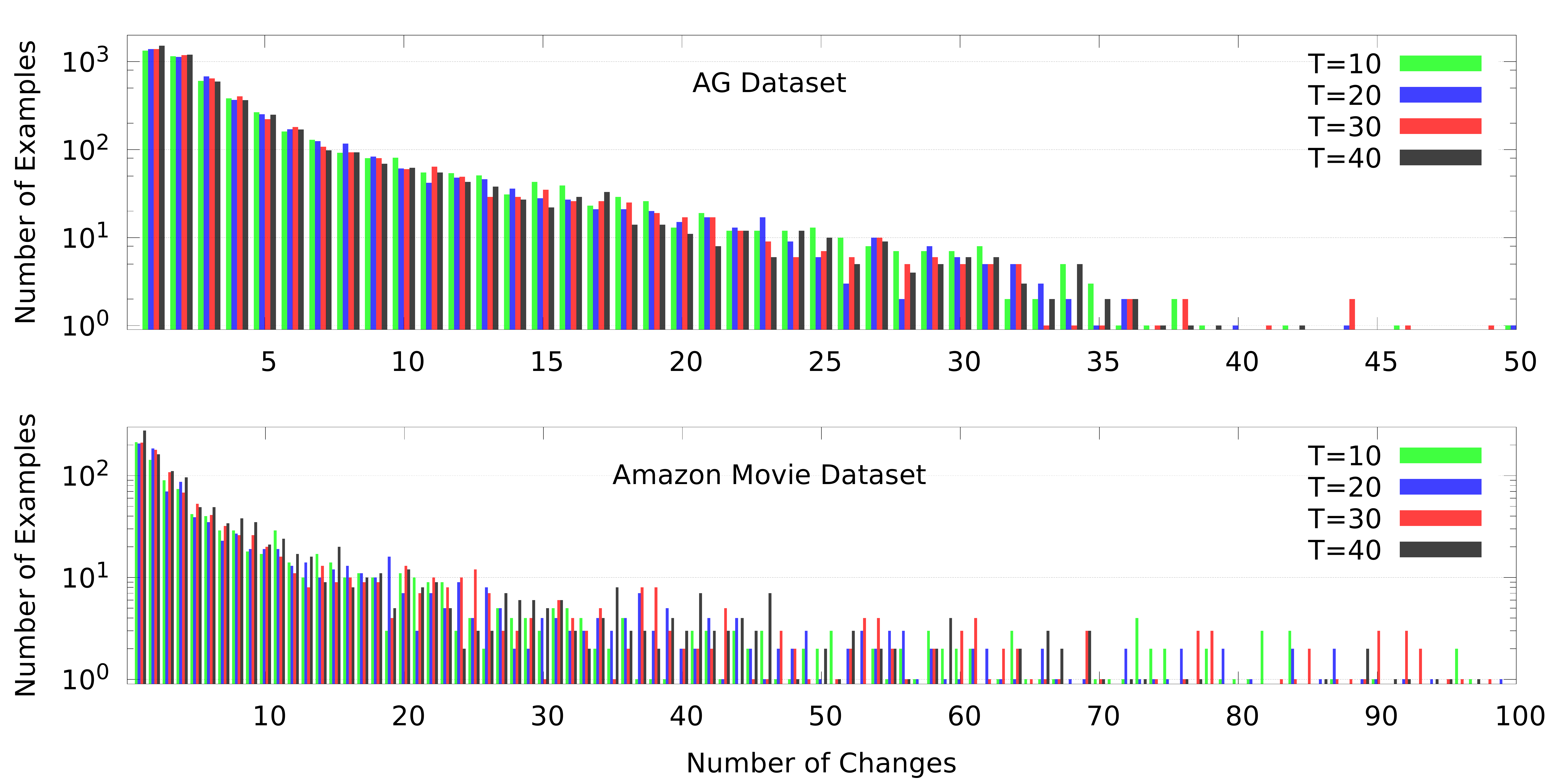}
  \caption{Distribution of changes for generated adversarial examples for networks trained with defensive distillation; note the logarithmic scale on the y-axis.}
  \label{fig:distillation:distribution}
 \end{figure}

 When looking at the number of changes (Tab.~\ref{tab:distillation:detail}), we can see that distillation makes it slightly more difficult to generate adversarial examples.
 The mean number of changes went up for all experiments (AG: $3.52$ to $3.94-4.47$, Amazon movie: $14.76$ to $17.31-19.29$).
 A similar increase can be seen in some instances for the median number of changes (Amazon movie: $3$ to $4$), however, these seem to be minor.
 The difference between the mean and the median can be explained if we look at the distribution in Fig.~\ref{fig:distillation:distribution}: while most generated adversarial examples only need a few changes, a small number of adversarial examples need a large number of changes increasing the mean value.

 Another aspect is robustness against transferability.
 Tab.~\ref{tab:distillation:transferability} shows that the transferability rate for the AG ($0.323-0.337$ compared to $0.369$) and Amazon movie ($0.199-0.253$ compared to $0.250$) datasets is a bit lower in most cases compared to the retrained network without distillation, with a difference of only $0.032-0.051$.
 Based on this, we conclude that defensive distillation has only a small impact on transferability and does not, in fact, prevent it.

\section{Discussion}
 Our results indicate that, at least on the two tested datasets, defensive distillation does not have the same effect for text classification as it has for image classification.
 The robustness against adversarial examples of networks trained with defensive distillation increases only slightly.
 One reason for this might be that defensive distillation effectively performs \textit{gradient masking} \cite{Tramer2017,Papernot2016_5}, which works against methods which directly or indirectly add the gradient to the input.
 However, in the algorithm used in our experiments the value of the gradient is only used to measure the importance of a given word on the network's final output, not directly added onto the input.
 The exact characteristics of the gradient are not important and, as a result, the \textit{gradient masking} itself has only a minimal effect on our algorithm.

 This hypothesis is further boosted by the results of Carlini and Wagner \cite{Carlini16}, who were able to generate adversarial examples for networks trained with defensive distillation with a slight modification to the generating algorithm on image classification tasks.
 They did this by restoring the gradient from the \textit{gradient masking} done by defensive distillation.
 This shows that the gradient still holds enough information for the generation of adversarial examples for image classification tasks, but it might not be as easily accessible as without defensive distillation.
 If this hypothesis is true, this might mean that other methods based on gradient masking are also not effective in the text domain.

\begin{table}[t]
 \centering
 \ra{1.2}
 \caption{\emph{Transferability} of generated adversarial examples to networks trained with defensive distillation with Temperature T.}
 \label{tab:distillation:transferability}

 \setlength{\tabcolsep}{1.5em}
 \begin{tabular}{@{}llccc@{}}
 \toprule
  T & Dataset & \makecell{Number\\ tested} & \makecell{Success\\ rate} & \makecell{Success rate \\w/o distillation} \\
  \midrule
  \multirow{2}{*}{10} & AG & 8150 & 0.323 & 0.369 \\
   & Amazon movie & 1157 & 0.218 & 0.250 \\ \midrule
  \multirow{2}{*}{20} & AG & 8126 & 0.331 & 0.369 \\
   & Amazon movie & 1217 & 0.199 & 0.250 \\ \midrule
  \multirow{2}{*}{30} & AG & 8141 & 0.325 & 0.369 \\
   & Amazon movie & 1229 & 0.253 & 0.250 \\ \midrule
  \multirow{2}{*}{40} & AG & 8134 & 0.337 & 0.369 \\
   & Amazon movie & 1195 & 0.211 & 0.250 \\
  \bottomrule
 \end{tabular}
\end{table}

\section{Conclusion}
 In this paper, we showed that, at least on the two tested datasets, defensive distillation does not work for hardening neural networks for text classification against adversarial examples.
 This still leaves two questions open: is gradient masking an option for increasing the robustness of neural networks and, if not, how can the robustness of neural networks for text classification be increased?
 To answer the first question, the effect of \textit{gradient masking} \cite{Tramer2017,Papernot2016_5} for both image classification and text classification should be studied in more detail, especially since similar results for another method of increasing robustness for image classification (saturated networks \cite{Nayebi2017}) were achieved by Brendel and Bethge \cite{Brendel2017}.
 For this, different methods of \textit{gradient masking} should be analyzed in future experiments to see if the gradient could be restored in a similar fashion.
 To answer the second question, different methods of hardening neural networks for text classification need to be examined.
 The question still remains open whether different methods from image classification can be successfully transferred to text classification or whether completely new approaches must be developed.

\subsubsection*{Acknowledgments.}
The authors gratefully acknowledge partial support from the German Research Foundation DFG under project CML (TRR 169) and the European Union under project SECURE (No 642667).
The following software libraries were used for this work: Keras, Tensorflow, Gensim, NLTK with the WordNet interface, and NumPy.

\bibliographystyle{splncs03doi} 
\bibliography{References}

\end{document}